\documentclass[10pt,twocolumn,letterpaper]{article}

\usepackage{ijcb}
\usepackage{times}
\usepackage{epsfig}
\usepackage{graphicx}
\usepackage{amsmath}
\usepackage{amssymb}

\usepackage{float}
\usepackage{subfig}
\usepackage[flushleft]{threeparttable}
\usepackage{adjustbox}
\usepackage{tablefootnote}
\usepackage[font={footnotesize}]{caption}
\usepackage{tablefootnote}
\usepackage{amsmath}
\usepackage{xcolor} 
\usepackage{dblfloatfix} 
\usepackage{caption}
\usepackage{booktabs,array}
\usepackage[detect-none]{siunitx}
\usepackage{multicol}
\usepackage[usestackEOL]{stackengine}
\usepackage{algorithmic}
\usepackage{algorithm}
\usepackage{hyperref}

\ijcbfinalcopy 

\ifijcbfinal\fi
\begin{document}

\title{White-Box Evaluation of Fingerprint Matchers: \\Robustness to Minutiae Perturbations}

\author{Steven A.~Grosz\\
Michigan State University\\
East Lansing, MI, 48824\\
{\tt\small groszste@msu.edu}
\and
Joshua J.~Engelsma\\
Michigan State University\\
East Lansing, MI, 48824\\
{\tt\small engelsm7@msu.edu}
\and
Nicholas G.~Paulter\\
National Institute of Standards and Technology\\
Gaithersburg, Maryland, 20899\\
{\tt\small paulter@nist.gov}
\and
Anil K.~Jain\\
Michigan State University\\
East Lansing, MI, 48824\\
{\tt\small jain@msu.edu}
}

\maketitle
\thispagestyle{empty}

\begin{abstract}
   Prevailing evaluations of fingerprint recognition systems have been performed as end-to-end black-box tests of fingerprint identification or authentication accuracy. However, performance of the end-to-end system is subject to errors arising in any of its constituent modules, including: fingerprint scanning, preprocessing, feature extraction, and matching. Conversely, white-box evaluations provide a more granular evaluation by studying the individual sub-components of a system. While a few studies have conducted stand-alone evaluations of the fingerprint reader and feature extraction modules of fingerprint recognition systems, little work has been devoted towards white-box evaluations of the fingerprint matching module. We report results of a controlled, white-box evaluation of one open-source and two commercial-off-the-shelf (COTS) minutiae-based matchers in terms of their robustness against controlled perturbations (random noise and non-linear distortions) introduced into the input minutiae feature sets. Our white-box evaluations reveal that the performance of fingerprint minutiae matchers are more susceptible to non-linear distortion and missing minutiae than spurious minutiae and small positional displacements of the minutiae locations.
\end{abstract}

\begin{figure}[!t]
\begin{center}
\includegraphics[width=0.48\textwidth]{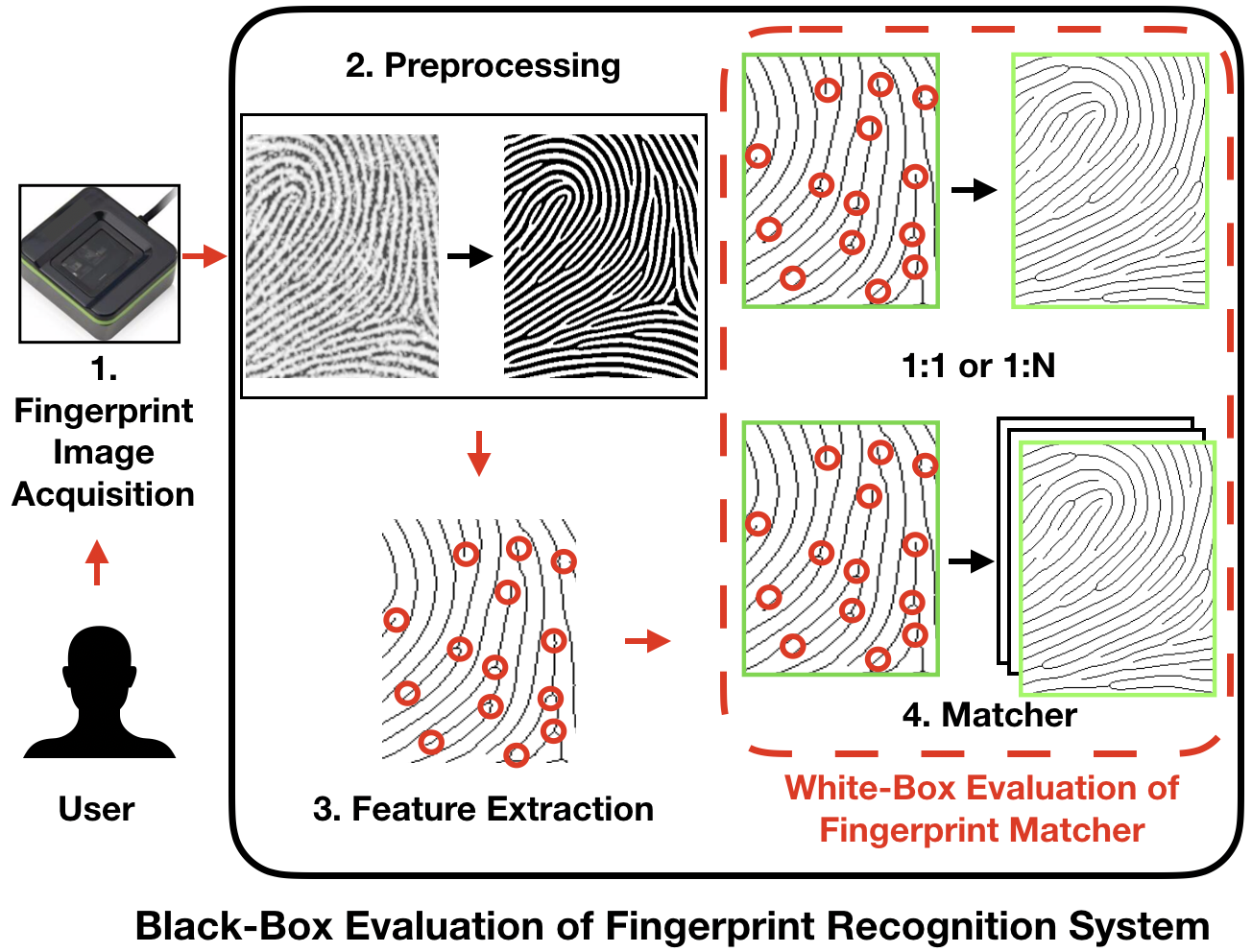}
\caption{Pipeline of a typical fingerprint recognition system: (1) fingerprint acquisition, (2) preprocessing, (3) feature extraction, and (4) matching. While existing evaluations of recognition systems primarily consist of an end-to-end black-box evaluation (i.e. all 4 modules simultaneously), we propose an independent, white-box evaluation of the matching module.}
\label{fig:fig1}
\end{center}
\vspace{-2.5em}
\end{figure} 

\section{Introduction}
The use of fingerprints in biometrics has seen a tremendous growth over the past 20 years due to their purported uniqueness, permanence, universality, and ease of collection \cite{handbook}. As fingerprint recognition technology continues to see widespread adoption, the need to understand and validate system recognition accuracy and robustness is paramount~\cite{unar2014review}. Current methods for evaluating automated fingerprint identification systems (AFIS) consist, almost entirely, of end-to-end black-box testing.\footnote{Black-box testing focuses on testing the end-to-end system using inputs and outputs \cite{testing}. In contrast, white-box testing evaluates the internal sub-components of a system.} For example, the National Institute of Standards and Technology (NIST) conducts fingerprint vendor technology evaluations (FpVTE)~\cite{nist} and the University of Bologna conducts fingerprint verification competitions (FVC) (\cite{fvc2002},~\cite{fvc2004},~\cite{dorizzi2009fingerprint}) to evaluate fingerprint recognition systems on their operational performance, as measured in terms of computational requirements and recognition accuracy. Additionally, Cappeli \textit{et al}. studied the effects of the operational quality of fingerprint scanners on recognition accuracy \cite{cappelli2008operational}. Black-box evaluations are valuable in that they allow for overall comparisons between recognition systems in terms of performance from the perspective of an end-user. However, this black-box testing approach is limited in that it provides limited information on which sub-module (image acquisition, preprocessing, feature extraction, or matching) of the AFIS is actually causing recognition failures (Figure~\ref{fig:fig1}).


\begin{figure}
\begin{center}
\includegraphics[scale=0.7]{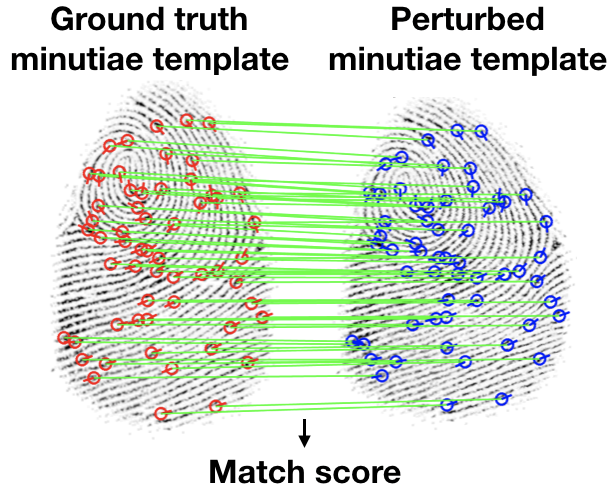}
\caption{Overview of the white-box testing of a minutiae-matcher. Input minutiae sets are perturbed by random positional perturbations and non-linear distortions. The perturbed minutiae sets are then matched to the unmodified templates to generate similarity scores. Lastly, a measurement uncertainty associated with each type of perturbation is computed, which indicates a measure of robustness to that type of perturbation.}
\label{fig:fig5}
\end{center}
\vspace{-2em}
\end{figure} 

To address this limitation inherent to black-box testing, we aim to build upon recent studies that performed white-box evaluations of individual modules of fingerprint recognition systems~\cite{3Dfingers, whole_hand, goldfinger, engelsma2018universal, chugh2017benchmarking}. To date, prior work in white-box evaluations of AFIS have primarily targeted the fingerprint reader and feature extraction modules. In particular,~\cite{3Dfingers, whole_hand, goldfinger, engelsma2018universal} designed and fabricated 3D fingerprint targets to evaluate fingerprint readers. Chugh \textit{et al}. conducted a white-box evaluation of minutiae extractors by examining their robustness to controlled levels of noise and motion blur~\cite{chugh2017benchmarking}. A few studies have also attempted a white-box evaluation of latent fingerprint examiners, by quantifying discrepancies between manual markups made by human experts~\cite{ulery2014measuring, ulery2015changes, hicklin2019gaze}. Furthermore, Peralta \textit{et al}. conducted a survey to compare the performance of top minutiae-based matching algorithms. Their work drew conclusions between different matching algorithms using various identification and authentication performance metrics on high quality fingerprint impressions~\cite{peralta2015survey}. Despite these efforts, they did not address the effects of varying fingerprint template (minutiae-set) fidelity (i.e., applying non-linear distortion and other random perturbations on the minutiae points) on scores output by the matching module.

It was shown in~\cite{chugh2017benchmarking} that the position ($x$, $y$, and $\theta$) and quantity of minutiae detected varies between different fingerprint minutiae extractors. However, the effects of these variations on minutiae-based matchers (MBMs) have yet to be studied. This work aims to evaluate the robustness of different MBMs to these slight perturbations of minutiae location, orientation, and quantity (additional spurious and missing minutiae). Additionally, non-linear distortion has been shown to affect the performance of fingerprint recognition systems and many methods have been proposed toward detecting and rectifying distortion in fingerprint images~\cite{ross2005fingerprint, chen2005fingerprint, ross2005deformable, ross2004estimating, dabouei2018fingerprint, dabouei2019deep, lan2019non}. Despite these efforts, non-linear distortion remains a significant challenge in fingerprint recognition~\cite{handbook}. Therefore, it is important that we quantify and compare the effects of non-linear distortion on various state-of-the-art fingerprint MBMs.

Given the lack of a standard, white-box evaluation of fingerprint matchers in the literature, we propose a rigorous, repeatable white-box evaluation protocol comprised of the following steps:

\begin{enumerate}
    \item Perturb a ``ground-truth" minutiae-set by (i) random translation and rotation shifts to the individual minutiae, (ii) random minutiae additions and deletions, (iii) combined random translation/rotation shifts and addition/removal of minutiae, and (iv) non-linear distortion (thin-plate spline distortion model~\cite{tps}).
    \item Compare unperturbed ``ground-truth" templates and the perturbed templates to obtain similarity score distributions associated with each perturbation.
    \item Compute an uncertainty measurement for each matcher due to the perturbation techniques.
\end{enumerate}

Our analysis aims to measure the robustness of various MBMs against the induced minutiae perturbations and reveal insights into the strengths of the MBMs, not previously revealed with black-box testing. More concretely, the contributions of this research are as follows:

\begin{enumerate}
    \item A detailed white-box evaluation protocol and uncertainty analysis of the fingerprint matching module. This evaluation augments previous studies on white-box evaluations of the fingerprint reader \cite{3Dfingers}\cite{whole_hand}\cite{goldfinger} and feature extractor modules \cite{chugh2017benchmarking}.
    \item An analysis on the effects of random positional perturbations and non-linear distortion on minutiae-matcher performance.
    \item Benchmarking two state-of-the-art commercial-off-the-shelf (COTS) MBMs\footnote{These two matchers are among the highest performing matchers in the NIST FpVTE.} and one open source MBM (SourceAFIS)~\cite{sourceAFIS} in accordance with our proposed protocol.
\end{enumerate}

\vspace{-.2cm}
\section{Evaluation Procedure}
We conduct two main experiments in our white-box evaluation. First, we perform an uncertainty analysis resulting from realistic amounts of non-linear distortion and random perturbation of minutiae positions. The positional perturbation parameters are taken from the experiment by Chugh \textit{et al.} of minutiae feature extractors on five live fingerprint impression databases: FVC 2002 (DB1A and DB3A), FVC 2004 (DB1A and DB3A) and NIST SD27, whereas the distortion parameters were estimated from a Thin-Plate-Spline (TPS) model learned from actual distorted fingerprints~\cite{chugh2017benchmarking}. Thus, these perturbation parameters can be considered representative of the amount of perturbation that may occur in a realistic operational setting. An uncertainty analysis of the matcher module provides a measure on the performance of the matcher in comparing ground-truth minutiae templates with templates perturbed with various positional variations and distortion. An overview of the analysis is shown in Figure \ref{fig:fig5}. In an additional experiment, we evaluate recognition performance of each MBMs on increasing levels of perturbation and distortion.

\subsection{Dataset for Experiments}
\begin{figure}[!t]
\centering
\subfloat[]{\includegraphics[width=0.13\textwidth]{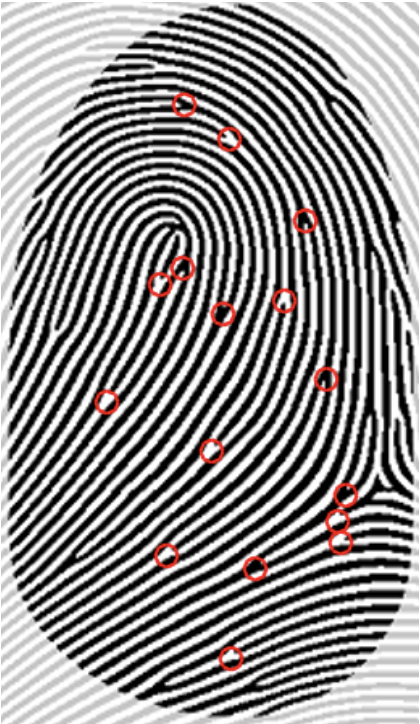}%
\label{print}}
\hfil
\subfloat[]{\includegraphics[width=0.13\textwidth]{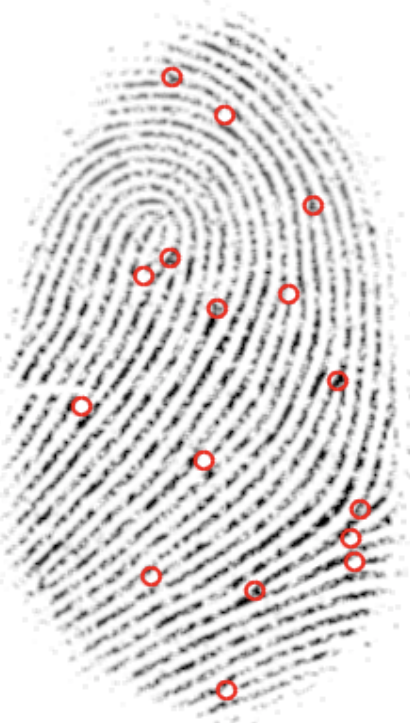}%
\label{print_minu}}
\hfil
\subfloat[]{\includegraphics[width=0.13\textwidth]{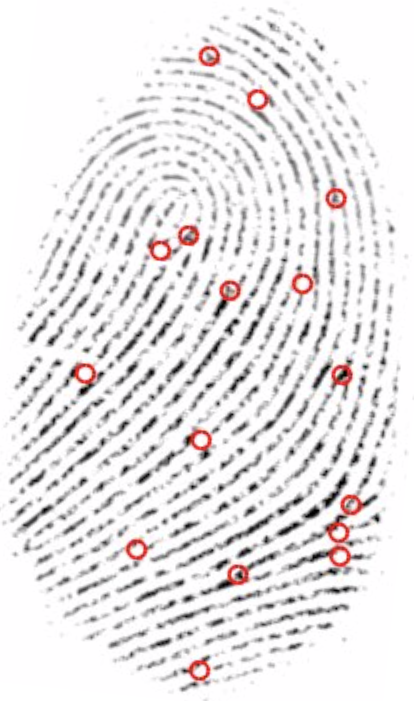}%
\label{print_minu2}}
\hfil
\caption{Example fingerprint images obtained from SFinGe synthetic fingerprint generator. (a) A master print with ground truth minutiae; (b) and (c) realistic impressions of the same fingerprint, derived from the master print.}
\label{fig:fig6}
\vspace{-1.5em}
\end{figure}

The first step in the white-box evaluation of MBMs is to obtain ground-truth minutiae-sets. We obtain the ground-truth minutiae-sets from fingerprints synthetically generated using SFinGe~\cite{sfinge}. We have generated 5000 unique ``master" reference fingerprints, each of which is used to produce two synthetic impressions that differ slightly in orientation, number of minutiae, and noise level (Fig.~\ref{fig:fig6}).\footnote{Here master refers to the original binary print generated by SFinGe prior to any distortion and/or noise that is introduced to produce realistic looking fingerprint images.} Thus, a total of 10 000 fingerprints are used (5000 unique fingers, 2 impressions per finger). For each of the 5000 master prints that SFinGe generates, a ground truth minutiae feature set is also produced. It is these minutiae sets that are fed as input to the fingerprint matchers for evaluation.

We utilize synthetic fingerprints because they provide reliable, consistent, ground-truth minutiae-sets. Otherwise, obtaining ground-truth minutiae-sets from an operational fingerprint dataset necessitates the use of minutiae sets extracted by the fingerprint recognition system's feature extractor on {\it noisy} fingerprints, such as that shown in Fig. \ref{fig:fig6} (b). This defeats the paramount goal of a white-box evaluation, which is to evaluate the performance of the matcher independent of the feature extractor (or other sub-modules) used.

\subsection{Perturbation Techniques}
\begin{figure*}[!t]
\centering
\subfloat[]{\includegraphics[width=0.16\textwidth]{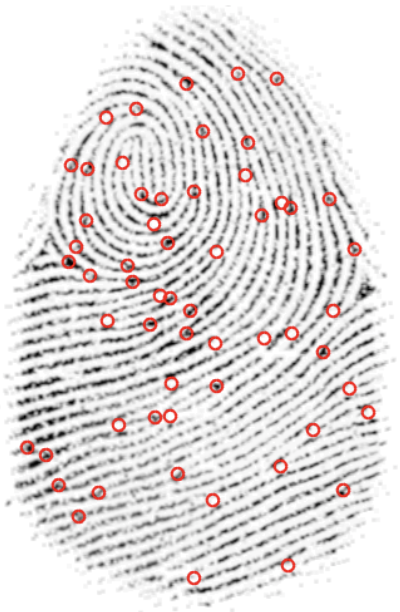}%
\label{case1}}
\hfil
\subfloat[]{\includegraphics[width=0.16\textwidth]{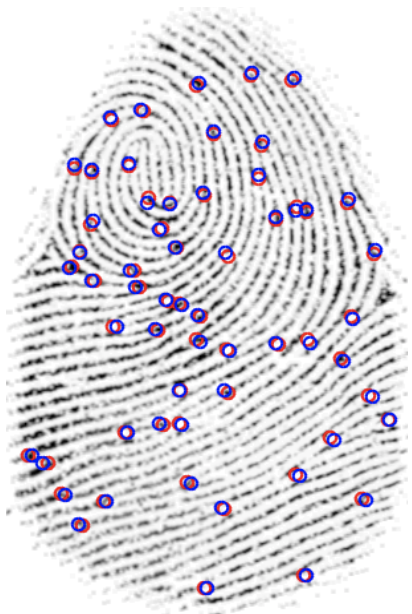}%
\label{case2}}
\hfil
\subfloat[]{\includegraphics[width=0.16\textwidth]{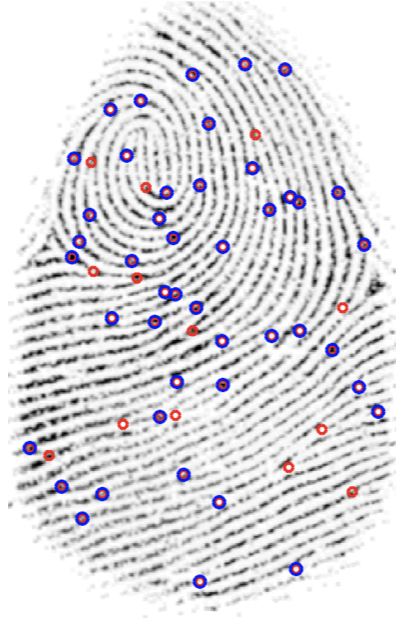}%
\label{case3}}
\hfil
\subfloat[]{\includegraphics[width=0.16\textwidth]{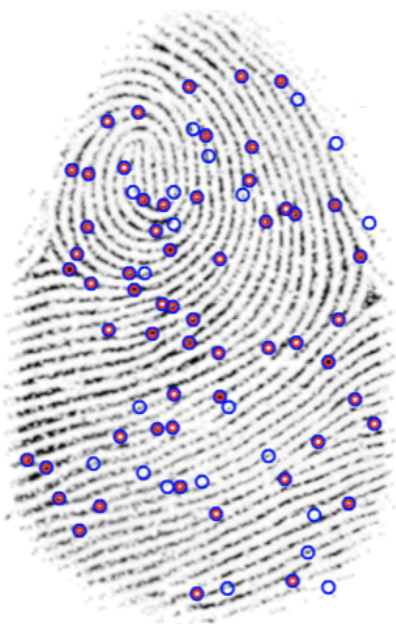}%
\label{case4}}
\hfil
\subfloat[]{\includegraphics[width=0.16\textwidth]{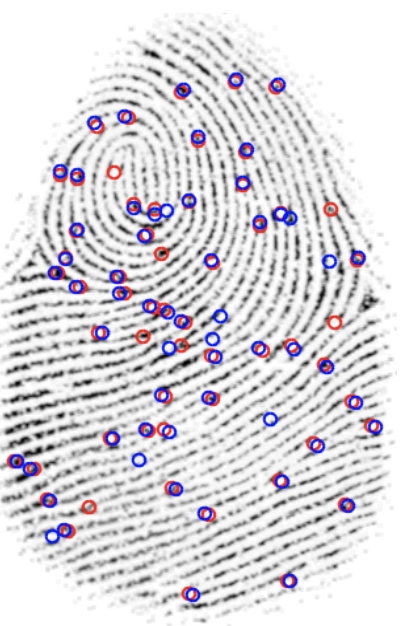}%
\label{case5}}
\hfil
\subfloat[]{\includegraphics[width=0.16\textwidth]{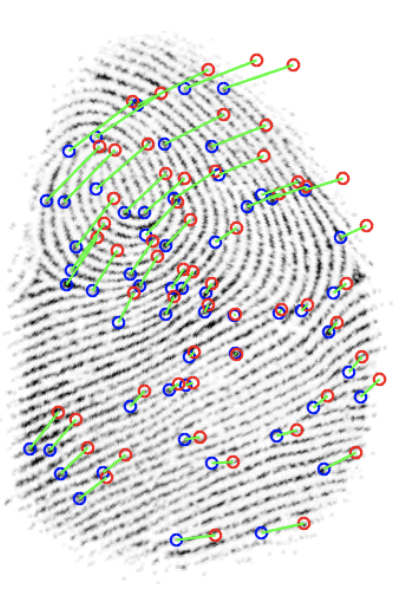}%
\label{case6}}
\caption{Illustration of the various minutiae perturbation techniques. For each image, the red circles denote the original minutiae locations and the blue circles designate the minutiae locations following the perturbations. (a) The original, ground truth minutiae locations, (b) random positional shifts of $x$, $y$, and $\theta$, (c) removal of random minutiae points, (d) addition of random minutiae points, (e) combined positional and missing/spurious minutiae, and (f) non-linear distortion of minutiae points.}
\label{perturbations}
\vspace{-1em}
\end{figure*}

Given a ground truth minutiae feature set of $n$ minutiae \{$(x_{1}, y_{1}, \theta_{1})$, ...,$(x_{n}, y_{n}, \theta_{n})$\}, the next step is to perform two types of perturbations: randomly generated positional perturbation and non-linear distortion. In particular, the minutiae perturbation techniques applied are: (i) random positional (translational and rotational) shifts, (ii) random addition and removal (spurious and missing) of minutiae, (iii) combined random positional shifts and addition/removal of minutiae, and (iv) non-linear distortion. These perturbation techniques are illustrated in Figure \ref{perturbations}.

\subsubsection{Translational and Rotational Shifts}
It was shown in \cite{chugh2017benchmarking} that different fingerprint minutiae extractors will provide slightly different $x$, $y$, and $\theta$ values. It is possible that these positional variations will have an impact on the fingerprint matcher performance. Therefore, the first perturbation technique involves random translation and rotation applied to the $x$, $y$, and $\theta$ of each minutiae. In a study on fingerprint uniqueness \cite{statistical_minu_models}, Zhu \textit{et al}. determined that a given minutiae point's position can be appropriately modeled by a multi-variate Gaussian distribution. Therefore, in keeping with established research, we use Gaussian distribution models to assign random noise to the positions and orientations of the fingerprint minutiae.

The mean, $\mu$, and standard deviation, $\sigma$, parameters of the Gaussian distributed translational ($\mu_p$ and $\sigma_p$) and rotational ($\mu_o$ and $\sigma_o$) variations are taken from the literature~\cite{chugh2017benchmarking}. In~\cite{chugh2017benchmarking}, a white-box evaluation of fingerprint feature extractors was performed on 3,458 real fingerprint images aggregated from five different public domain databases: FVC 2002 (DB1A and DB3A), FVC 2004 (DB1A and DB3A), and the NIST SD27 rolled prints database. In particular, the average translation and rotation errors were obtained as the distance between the minutiae locations provided by the automated feature extractor and the manually annotated ground truth minutiae locations. These values were binned (into 5 bins) based upon the NIST Fingerprint Image Quality (NIFQ) 2.0 \cite{NFIQ2} metric associated with the input fingerprint to the feature extractor.\footnote{The values assigned by the NFIQ 2.0 quality metric range from 0 to 100 and the ranges of the individual bins are [0, 20], [21, 40], [41, 60], [61, 80], and [81, 100].} Therefore, we take the weighted average of these transformation parameters across each bin (weights based on number of images in the bin) as the mean and standard deviation for our models of Gaussian distributed perturbation (Table~\ref{params}).
\begin{table}
\renewcommand{\arraystretch}{1.3}
\caption{Parameter Values for Realistic Perturbations}
\label{params}
\centering
\small 
{
\begin{tabular}{|>{\flushleft}m{0.3\textwidth}||m{0.12\textwidth}|}
\hline
Perturbation Technique & Parameters\\
\hline
\hline
\textbf{Positional Displacement} \newline (Pixels, radians) & $\mu_p = 4.048$ \newline $\sigma_p = 0.688$ \newline $\mu_o = 0.130$ \newline $\sigma_o = 0.071$\\
\hline
\textbf{Missing Minutiae} \newline (\#Missing / \#Ground Truth) & $\mu_m = 0.209$ \newline $\sigma_m = 0.106$\\
\hline
\textbf{Spurious Minutiae} \newline (\#Spurious / \#Ground Truth) & $\mu_s = 0.523$ \newline $\sigma_s = 0.349$\\
\hline
\textbf{Non-Linear Distortion} \newline (Eigenvector coefficient $c_i$) & $\mu_d = 0.0$ \newline $\sigma_d = 0.66$\\
\hline
\end{tabular}
}
\vspace{-1.6em}
\end{table}


\subsubsection{Missing and Spurious Minutiae}
A commonly occurring variation in minutiae feature sets is the number of minutiae points detected in a fingerprint image by various feature extractors. The number of detected minutiae points can either be lower than ground-truth (due to missed detections) or higher than ground-truth (false detections). Therefore, our evaluation protocol takes into account the effects of spurious and missing minutiae on MBMs.

Similar to the previously discussed positional perturbation technique, the distributions of missing minutiae and spurious minutiae are modeled by Gaussian distributions whose parameters are set as the expected value and variance of the differences observed in the output of various feature extractors~\cite{chugh2017benchmarking}. As in \cite{chugh2017benchmarking}, the mean and standard deviation values are reported as a ratio of the number of missed minutiae to the total number of ground truth minutiae in the fingerprint image (Table~\ref{params}).


\begin{figure}
\centering
\includegraphics[width=0.47\textwidth]{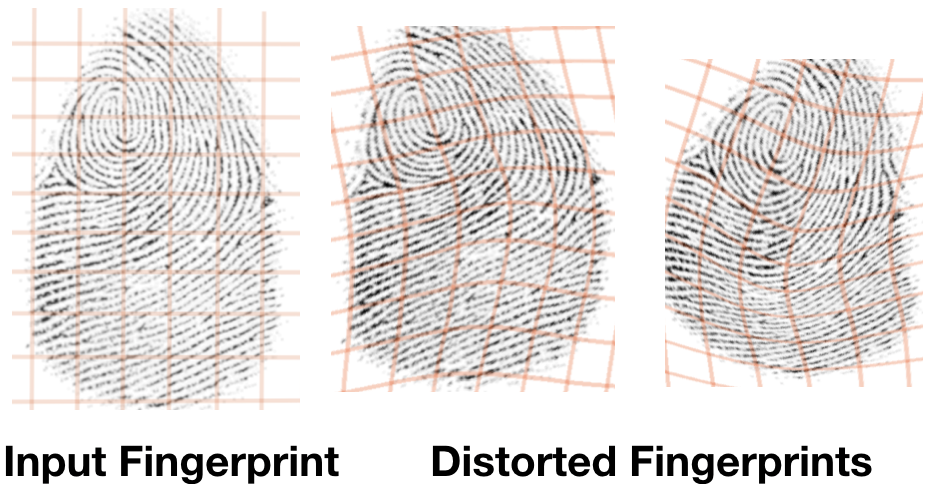}%
\caption{Applying non-linear distortion fields (obtained from a statistical model~\cite{nonlinear_distortion}) to an input fingerprint. Note, we show the fingerprint images here to illustrate the
effects of the non-linear distortion on fingerprint images. However, in the experiments, we directly apply the distortion field to
ground truth minutiae points provided by SFinge, eliminating the need of a feature extractor for this white-box evaluation of the matcher.}
\label{fig:distortion}
\vspace{-1.5em}
\end{figure}

\subsubsection{Combined Random Perturbation}
This perturbation technique involves aggregating the previously discussed techniques into one. This technique is intended to show the combined effects of the positional perturbations and the presence of missing and spurious minutiae, which is more representative of what a matching module might reasonably be expected to receive from a feature extractor.

First, we delete some minutiae points from the input minutiae feature set as described previously, followed by insertion of some spurious minutiae. Lastly, we apply the translation and rotation perturbations. The missing minutiae perturbation technique is first applied so that none of the added spurious minutiae are deleted.

\subsubsection{Non-Linear Distortion Model}
The final perturbation technique applied to the input minutiae feature set is that of an applied non-linear distortion model proposed by Si, \textit{et al}. in~\cite{nonlinear_distortion}. We chose to apply a non-linear distortion model (learned from actual distorted fingerprints) as a perturbation since it is well known that much of the perturbation of the minutiae points in an operational scenario will be from distortion introduced by the elastic friction ridge pattern on the human fingertip, as observed in the work of Gutierrez da Costa, \textit{et al}. \cite{da2014evaluation}. Furthermore, it is important to evaluate the robustness of MBMs against non-linear distortion since it has been demonstrated that hackers can obfuscate their own identity by intentionally introducing a heavy distortion to their fingerprints during the fingerprint acquisition process~\cite{nonlinear_distortion}. 

The non-linear distortion model proposed in \cite{nonlinear_distortion} was formulated and trained as follows. (i) A database of 320 distorted fingerprint videos was acquired. (ii) The minutiae were then extracted from the first and last frame of the video sequence. (iii) Correspondences were computed between the minutiae from the first frame and the minutiae in the final, distorted fingerprint frame. (iv) The corresponding minutiae points were used to estimate a distortion field $d_i$ (grid) using a TPS model~\cite{tps}, which is commonly used to model fingerprint distortion~\cite{ross2004estimating, ross2005fingerprint, dabouei2018fingerprint, bazen2002thin, bazen2002elastic, zhang2007fake, bazen2003fingerprint}. (v) Each distortion field $d_i$ was flattened into a vector and used to compute a mean distortion field $\bar{d}$. Subsequently, a covariance matrix $D$ was constructed. (vi) From the covariance matrix, Principle Component Analysis (PCA) was used as a statistical model to capture the variance of the training distortion fields. Therefore, a new distortion field $d$ can be generated from a subset of $t$ eigenvectors in accordance with Equation~\ref{eq:dist_field}:

\begin{equation}
    d \approx \bar{d}~\sum_{i = 1}^{t}c_i\sqrt{\lambda_ie_i}
    \label{eq:dist_field}
\end{equation} where $c_i$ are the coefficients of the eigenvectors $e_i$.

In our work, we select the coefficients $c_i$ of two eigenvectors (with the largest two eigenvalues) to generate new distortion fields from a normal distribution with mean of 0 and standard deviation of $\frac{2}{3}$. Unlike in \cite{nonlinear_distortion}, these randomly generated non-linear distortion fields are then applied directly to an unperturbed minutiae set rather than the fingerprint image. We directly obtain the new $x$ and $y$ location of each minutiae point by following where these coordinates end up in the newly distorted grid, whereas the new orientation is aligned in the direction of the line connecting this point and the coordinate which is ten pixels away along the distorted ridge. Examples of fingerprints before and after applying the non-linear distortion model are shown in Figure~\ref{fig:distortion}.

\newcolumntype{M}[1]{>{$\centering\arraybackslash}m{#1}<{$}}

\begin{table*}
\caption{Increasing Perturbation Parameters}
\label{roc_parameters}
\centering
\begin{tabular}{|>{\flushleft}m{0.3\textwidth}||M{0.05\textwidth}|M{0.05\textwidth}|M{0.05\textwidth}|M{0.05\textwidth}|M{0.05\textwidth}|M{0.05\textwidth}|M{0.05\textwidth}|M{0.05\textwidth}|}
\hline
Iteration Number & \mathbf{1} & \mathbf{2} & \mathbf{3} & \mathbf{4} & \mathbf{5} & \mathbf{6} & \mathbf{7} & \mathbf{8}\\
\hline
\hline
\textbf{Positional Displacement} \newline ($\begin{vmatrix}\Delta x + \Delta y\end{vmatrix}$, \newline $[\mu_{o}, \sigma_{o}]$) & 5, \newline [0,5] & 10, \newline [0,10] & 15, \newline [0,15] & 20, \newline [0,20] & 25, \newline [0,25] & 30, \newline [0,30] & 35, \newline [0,35] & 40, \newline [0,40]\\
\hline
\textbf{Missing Minutiae} \newline (Missing Minutiae / Ground Truth) & 0.20 & 0.40 & 0.60 & 0.80 & 0.85 & 0.90 & 0.95 & 0.99\\
\hline
\textbf{Spurious Minutiae} \newline (Spurious Minutiae / Ground Truth) & 0.5 & 1.0 & 1.5 & 2.0 & max & max & max & max\\
\hline
\textbf{Combined Random Perturbation} \newline ($\begin{vmatrix}\Delta x + \Delta y\end{vmatrix}$, \newline $[\mu_{o}, \sigma_{o}]$, \newline Missing Minutiae / Ground Truth, \newline Spurious Minutiae / Ground Truth) & 5, \newline [0,5] \newline 0.20, \newline 0.5 & 10, \newline [0,10] \newline 0.40, \newline 1.0 & 15, \newline [0,15] \newline 0.60, \newline 2.0 & 20, \newline [0,20] \newline 0.80, \newline 1.5 & 25, \newline [0,25] \newline 0.85, \newline max & 30, \newline [0,30] \newline 0.90, \newline max & 35, \newline [0,35] \newline 0.95, \newline max & 40, \newline [0,40] \newline 0.99, \newline max\\
\hline
\textbf{Non-Linear Distortion} \newline ($\sigma$ of eigenvector coefficients $c_i$ with $\mu$ = 0) & 0.5 & 1.0 & 1.5 & 2.0 & 2.5 & 3.0 & 3.5 & 4.0\\
\hline
\multicolumn{9}{c}{Max refers to the maximum number of minutiae points (255) as allowed by the ISO/IEC 19794-2~\cite{ISO_19794_2} standard.}
\end{tabular}
\vspace{-1.5em}
\end{table*}

\subsection{Uncertainty Analysis}
We evaluate matcher performance on each of the given perturbation techniques. Similarity scores between the original fingerprint template and the perturbed template are computed. From these scores, a total uncertainty is assigned to each fingerprint matcher chosen for the study. The step-by-step procedure used to calculate the uncertainty of a fingerprint matcher to the controlled, realistic perturbations (random positional perturbations and non-linear distortions) is as follows:

\begin{enumerate}
    \item Generate $M = 10 000$ $A_{ref,k}$ reference fingerprint impressions ($2$ impressions from each $5000$ unique master prints), $1 \leq k \leq M$.
    \item Obtain $M$ minutiae feature sets, $S_{ref,k}$, from each $A_{ref,k}$.
    \item For each $S_{ref,k}$, synthesize $N = 100$ perturbed minutiae sets, $S'_{test,k,n}$, $1 \leq n \leq N$. 
    \item Generate genuine similarity scores, $s_{k,n}$, between $S_{ref,k}$ and each $S'_{test,k,n}$ using the public-domain minutiae matcher~\cite{sourceAFIS}, COTS-A, and COTS-B.
    \item Normalize the scores, $s_{k,n}$, to be in the range of $[0,1]$ using min-max normalization, where the min and max are matcher specific values.
    \item Compute the average, $\mu_{k}$, of the $s_{k,n}$ scores using $\mu_{k} = \frac{1}{N}\sum_{n=1}^{N}(s_{k,n})$
    \item Compute the standard uncertainty, $u_k$, of $A_{k}$ using $u_{k} = \sqrt{\frac{1}{N}\sum_{n=1}^{N}(\mu_{k}-s_{k,n})^{2}}$
    \item Repeat steps $1$ to $7$ for each reference feature set, obtaining a $u_k$ for each $S_{ref,k}$.
    \item Compute the total uncertainty, $u_{matcher}$, for the matcher using $u_{matcher} = \sqrt{\frac{1}{M}\sum_{k=1}^{M}u_k^2}$
\end{enumerate}

This uncertainty calculation uses the Monte Carlo method~\cite{jcgm} for estimating uncertainties. It demonstrates the sensitivity of a matcher to perturbations of its inputs. A lower uncertainty score is better as it indicates better robustness to the perturbations. To augment the findings of the uncertainty calculation, we also compute genuine and imposter similarity scores on the perturbed templates. In other words, we obtain a distribution of genuine similarity scores from corresponding (unperturbed, perturbed) impressions and a distribution of imposter scores from non-corresponding (unperturbed, perturbed) impressions to evaluate the effects of the perturbation on recognition accuracy.

\subsection{Recognition Accuracy vs. Perturbation}
Here we discuss the experimental design for evaluating the recognition accuracy of the MBMs on increasing levels of perturbation. For this experiment, we select 2000 fingerprint impression pairs from our synthetically generated master prints. The experiment consists of running eight iterations at increasing levels of perturbation of each type (random positional shifts, random addition and removal of minutiae, combined random positional shifts and addition/removal of minutiae, and non-linear distortion) and obtaining 2000 genuine and 50 000 imposter similarity scores at each iteration~(Table~\ref{roc_parameters}). Additionally, for each perturbation type, the true acceptance rate (TAR) at a constant false acceptance rate (FAR) of $0.01 \%$ is obtained at each iteration to track the recognition accuracy vs. perturbation amount. 

\section{Experimental Results}
\begin{figure*}
\centering
\begin{tabular}{|p{0.015\textwidth}|m{0.29\textwidth}|m{0.29\textwidth}|m{0.29\textwidth}|}
\hline
 & COTS-A & COTS-B & SourceAFIS \\
\hline
\rotatebox[origin=c]{90}{Combined Random Perturbation } & \subfloat[]{\includegraphics[width=0.29\textwidth]{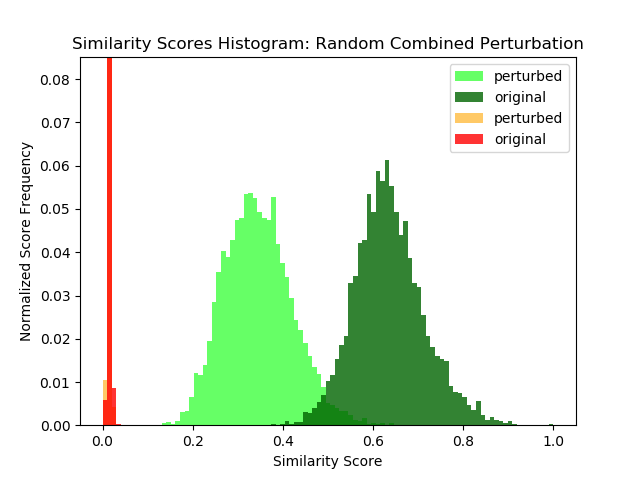}%
\label{Imposter Match Scores Histogram: Combined COTS-A}} & \subfloat[]{\includegraphics[width=0.29\textwidth]{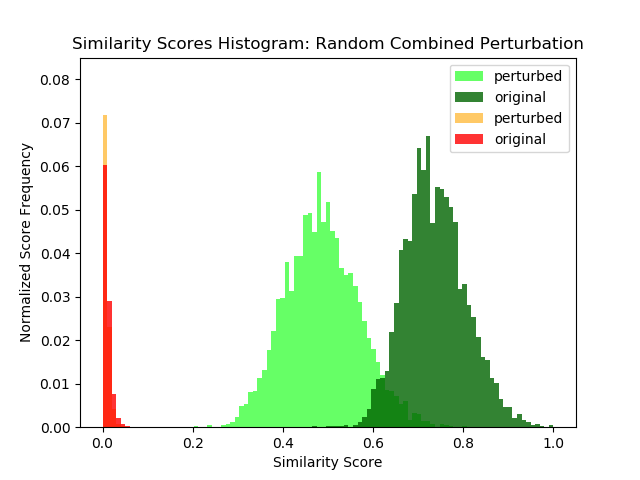}%
\label{Imposter Match Scores Histogram: Combined COTS-B}} & \subfloat[]{\includegraphics[width=0.29\textwidth]{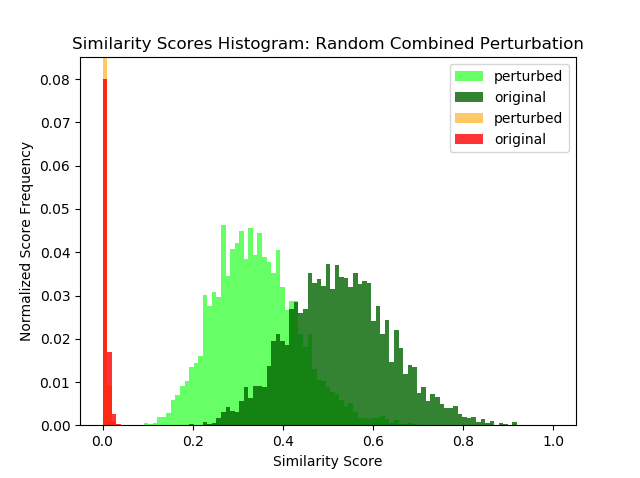}%
\label{Imposter Match Scores Histogram: Combined SourceAFIS}} \\
\hline
\rotatebox[origin=c]{90}{Non-Linear Distortion} & \subfloat[]{\includegraphics[width=0.29\textwidth]{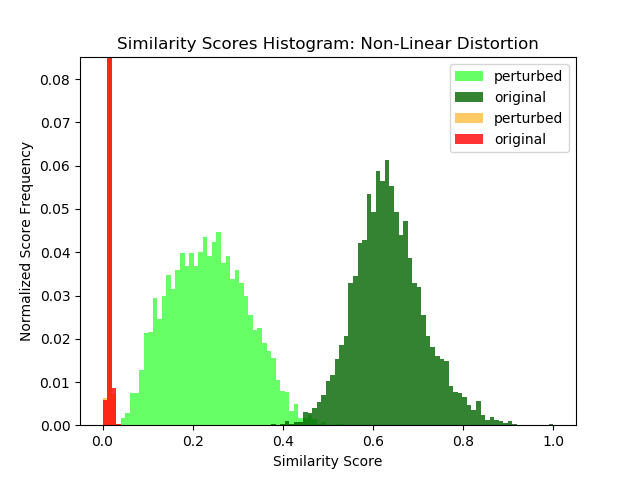}%
\label{Imposter Match Scores Histogram: Distorted COTS-A}} & \subfloat[]{\includegraphics[width=0.29\textwidth]{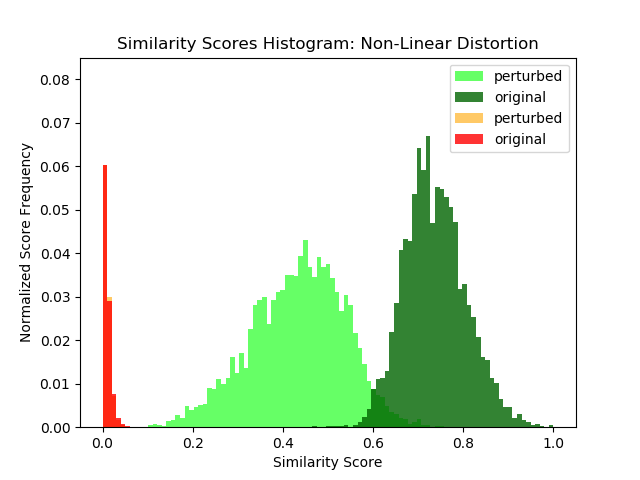}%
\label{Imposter Match Scores Histogram: Distorted COTS-B}} & \subfloat[]{\includegraphics[width=0.29\textwidth]{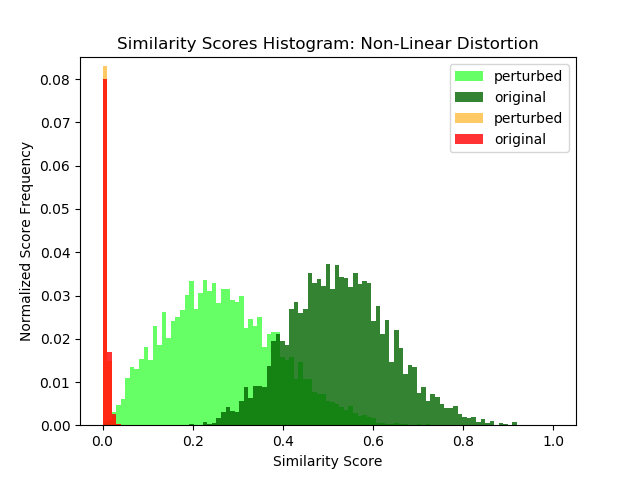}%
\label{Imposter Match Scores Histogram: Distorted SourceAFIS}} \\
\hline
\end{tabular}
\caption{Comparison on the effects of the combined random perturbations (translation/rotation shifts, missing minutiae, and spurious minutiae) and non-linear distortion on the genuine and imposter similarity scores obtained from each minutiae matcher: COTS-A, COTS-B, and SourceAFIS. The green and light-green distributions are the unperturbed and perturbed genuine similarity scores, respectively. The red and orange distributions are the unperturbed and perturbed imposter scores (scaled by 0.1), respectively.}
\label{fig:histograms}
\vspace{-1em}
\end{figure*}

\begin{figure*}[!hb]
\centering
\subfloat[]{\includegraphics[width=0.33\textwidth]{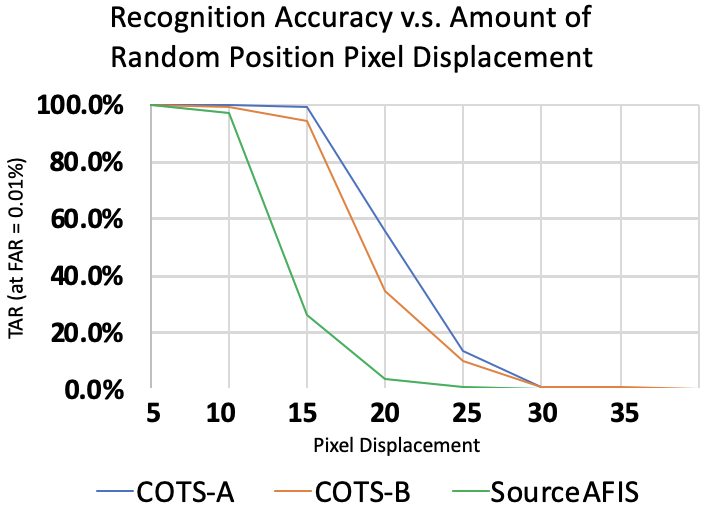}%
\label{1case}}
\hfil
\subfloat[]{\includegraphics[width=0.33\textwidth]{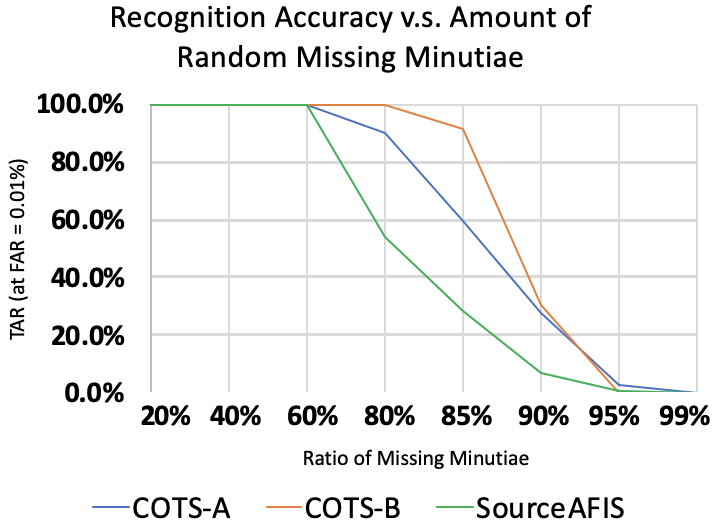}%
\label{2case}}
\hfil
\subfloat[]{\includegraphics[width=0.33\textwidth]{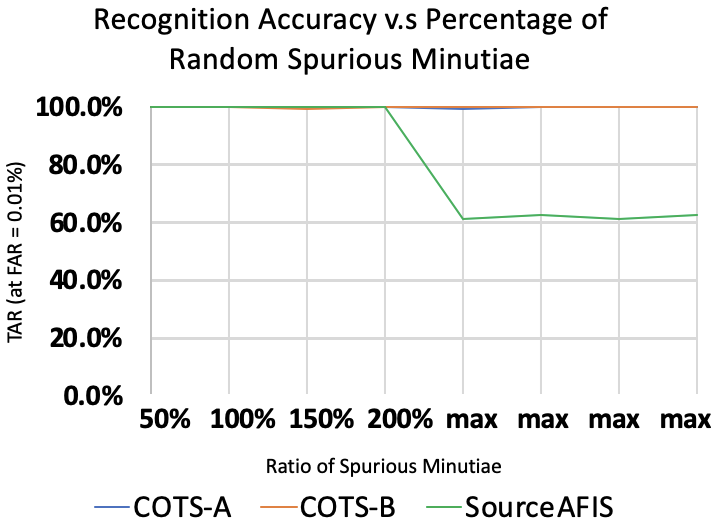}%
\label{3case}}
\hfil
\subfloat[]{\includegraphics[width=0.33\textwidth]{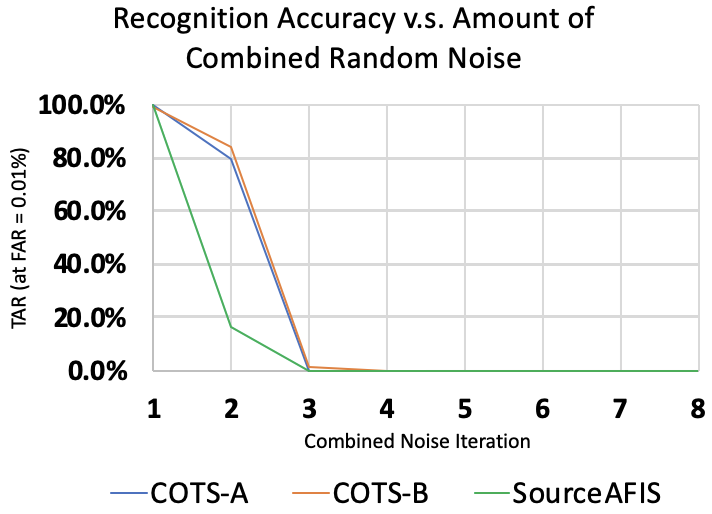}%
\label{4case}}
\hfil
\subfloat[]{\includegraphics[width=0.33\textwidth]{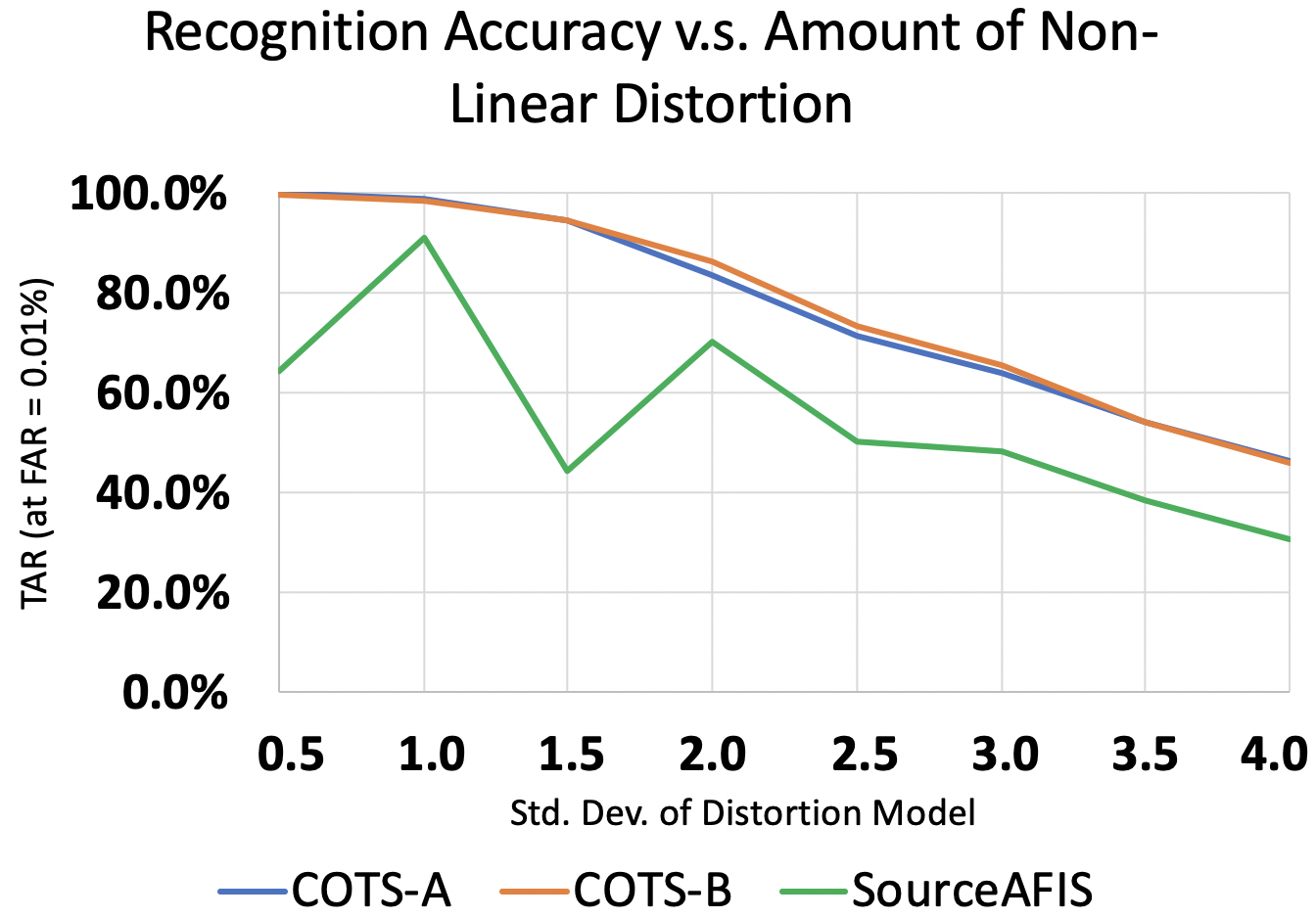}%
\label{5case}}
\caption{True accept rate at a constant false accept rate of $0.01\%$ of the three minutiae matchers vs. increasing levels of perturbation. (a) Increasing random minutiae position displacements (in pixels), (b) increasing percentage of ground truth minutiae removed, (c) increasing percentage of ground truth minutiae added, (d) increasing combined random noise (combined pixel displacements, missing minutiae, and spurious minutiae), and (e) increasing std. dev. values of the non-linear distortion model.}
\label{roc}
\end{figure*}

Using the synthetic fingerprint data set and perturbations defined previously, we evaluate the effects of the various perturbation techniques on the performance of each MBM. We examine the distributions of genuine and imposter similarity scores before and after the realistic perturbations and calculate the uncertainty observed in each of the MBMs. Histograms of the genuine and imposter similarity distributions for each MBM following the modeled perturbations are shown in Figure \ref{fig:histograms}. The measurement uncertainties associated with each type of perturbation are shown in Table \ref{uncertainty}. Finally, matcher recognition accuracy vs. increasing levels of perturbation and distortion are shown in Figure \ref{roc}.

\subsection{Effect of Random Perturbations}

To analyze the effects of random perturbation (position shifts and addition and removal of minutiae points), we compare the genuine and imposter similarity scores of each MBM before and after the applied perturbations. Overall, we see that the mean of the perturbed genuine score distributions are all significantly lower in value than the mean of the unperturbed distributions. Among the random perturbations, the shift in the mean is most pronounced due to the combined perturbation, followed by random removal of minutiae. Lastly, we note that there is a minimal impact to the imposter score distributions. It is not surprising that the imposter scores are minimally impacted since MBMs are fine-tuned for very low FAR, and thus have very peaked imposter distributions centered around 0.

In general, we notice that the variance of the genuine similarity distributions increases following the random perturbations. For each matcher, the largest increase in variance is seen with randomly removing minutiae, suggesting that the effect of missing minutiae has a more unpredictable and volatile impact on the performance of MBMs. This result agrees with a previous study that concluded that missing minutiae has a greater effect than spurious minutiae on the Delaunay triangulation method for fingerprint authentication~\cite{yang2014effect}. One explanation may be that certain minutiae contribute significantly more to the overall similarity score than others. Hence, the removal of such significant landmark points could drastically effect the similarity score produced by a MBM.

Analysis of the uncertainty scores for each matcher due to the various random perturbation techniques shown in Table~\ref{uncertainty} indicates that missing minutiae and the combined perturbation model generate the greatest measurement uncertainty. This agrees with the similarity score distributions where we observe the largest increase in variance due to the missing and combined perturbations.

\begin{table}
\renewcommand{\arraystretch}{1.3}
\caption{Uncertainty Scores of Various Realistic Perturbations}
\label{uncertainty}
\centering
\begin{tabular}{|m{0.09\textwidth}|>{\centering\arraybackslash}m{0.08\textwidth}|>{\centering\arraybackslash}m{0.08\textwidth}|>{\centering\arraybackslash}m{0.1\textwidth}|}
\hline
 & COTS-A & COTS-B & SourceAFIS \\
\hline
Positional & 0.001 & 0.001 & 0.001 \\
\hline
Missing & 0.009 & 0.007 & 0.005 \\
\hline
Spurious & 0.005 & 0.005 & 0.002 \\
\hline
Combined & 0.006 & 0.008 & 0.007 \\
\hline
Distorted & 0.029 & 0.025 & 0.028 \\
\hline
\end{tabular}
\vspace{-1.6em}
\end{table}

\subsection{Effect of Non-Linear Distortion}
Figure \ref{fig:histograms} (d), (e), and (f) show the effects of the non-linear distortion. Out of all perturbation techniques in the study, the non-linear distortion has the greatest impact on decreasing the mean of the genuine similarity score and increasing the variance for each matcher. The non-linear distortion also produces the greatest measurement uncertainty for each matcher. Thus, it is evident that realistic levels of non-linear distortion continue to have a tremendous effect on lowering the effectiveness of current MBMs. This motivates further research into MBMs or other fingerprint matching algorithms that are more robust to non-linear distortion, such as the recent deep learning approach taken in \cite{engelsma2019learning} which outperforms state-of-the-art minutiae based matchers on the distorted FVC 2004 DB1 A dataset. Alternatively, algorithms for detecting and correcting for non-linear distortion should be developed and incorporated into MBMs.

\subsection{Accuracy vs. Perturbation Amount}
Figure \ref{roc} shows the TAR at a constant FAR of $0.01 \%$ for each MBM. In panel (c) of Figure \ref{roc}, we observe that randomly adding spurious minutiae on its own has no effect on the recognition accuracy of the COTS MBMs and very little effect for the SourceAFIS matcher until exceeding $200 \%$ spurious minutiae. Furthermore, increasing the positional perturbation and the number of missing minutiae has very little effect for all matchers up to a perturbation level at which the performance drops steeply. Panels (a) and (b) of Figure \ref{roc} indicate that these inflection points are at about $15$ pixels to $20$ pixels (for $416 x 560$ images) for the positional displacement and around $80 \%$ to $85 \%$ for the percentage of missing ground truth minutiae. Panel (d) shows a similar sudden drop in recognition accuracy for the combined random perturbation around a combined 10 pixel displacement, $40 \%$ minutiae removal, and $100 \%$ minutiae addition. Interesting to note that individually, these levels of perturbation allow for greater than $95 \%$ accuracy, but when aggregated together produce near $0 \%$ accuracy. Finally, (e) shows that the effect on increasing amounts of non-linear distortion is more gradual, yet still detrimental to recognition accuracy.

Comparing the robustness of each matcher to the increasing perturbation, we see that the SourceAFIS matcher consistently shows the greatest decline in recognition accuracy. The degradation in recognition performance for the SourceAFIS matcher is also more steep compared to the other two matchers. This suggests that the COTS MBMs are relatively more robust to the induced perturbations, however, we see that with enough random positional perturbations and non-linear distortion, these matchers also show significant performance decline.

\section{Conclusion and Future Work}
We have presented a standardized evaluation protocol for white-box evaluation of fingerprint MBMs. In particular, we have computed the measurement uncertainty of an open source and two state-of-the-art COTS MBMs to various normally-distributed random perturbations and non-linear distortion of input minutiae feature sets. Our results indicate that missing minutiae and non-linear distortion to input minutiae locations may drastically affect the similarity scores output by MBMs. Specifically, it was found that non-linear distortion has the greatest effect of lowering the similarity scores between an unperturbed impression and a corresponding perturbed impression, whereas adding spurious minutiae had the least effect. These findings suggest that MBMs are relatively robust to slight positional and spurious minutiae perturbations, but weak to non-linear distortion and random removal of minutiae. This work could be extended to evaluate additional MBMs, such as Minutia Cylinder Code (MCC)~\cite{cappelli2010minutia}. Further study may evaluate the effect of correlated noise to account for non-uniformity caused by the elasticity of the human skin and the effect of fingerprint class (arch, whorl, right loop, and left loop) on matcher uncertainty, since it has been observed that minutiae distributions are class dependent~\cite{ross2007template}.

\section*{Acknowledgment}
This research was supported by grant no. 70NANB17H027 from the NIST Measurement Science program. The views and conclusions contained herein are those of the authors and should not be interpreted as necessarily representing the official policies, either expressed or implied, of NIST, or the U.S. Government. The U.S. Government is authorized to reproduce and distribute reprints for governmental purposes notwithstanding any copyright annotation therein.

{\small
\bibliographystyle{ieee}
\bibliography{citations}
}

\end{document}